# Firing Cell: An Artificial Neuron with a Simulation of Long-Term-Potentiation-Related Memory

Jacek Bialowas Dept. of Anatomy and Neurobiology Medical University of Gdansk Debinki 1, 80-211 Gdansk, Poland

Beata Grzyb Faculty of Mathematics, Physics and Computer Science Maria Curie-Sklodowska University Pl. Marii Curie-Skodowskiej 1 20-031 Lublin, Poland E-mail: beata@gabri.pl

Pawel Poszumski Institute of Oceanology Polish Academy of Science Powstancow Warszawy 55 81-712 Sopot, Poland

**Abstract:** We propose a computational model of neuron, called firing cell (FC), properties of which cover such phenomena as attenuation of receptors for external stimuli, delay and decay of postsynaptic potentials, modification of internal weights due to propagation of postsynaptic potentials through the dendrite, modification of properties of the analog memory for each input due to a pattern of short-time synaptic potentiation or long-time synaptic potentiation (LTP), output-spike generation when the sum of all inputs exceeds a threshold, and refraction. The cell may take one of the three forms: excitatory, inhibitory, and receptory. The computer simulations showed that, depending on the phase of input signals, the artificial neuron's output frequency may demonstrate various chaotic behaviors.

## 1. Introduction

The paper presents a new approach to building computational models of selected brain circuits, aimed to develop a tool supporting an interpretation of recordings of electric potentials in living neural tissue. Such computational models must reflect the most essential properties of biological neurons and their assemblies, but, in order to not to increase the necessary computational power beyond necessity, not too much more. We propose that, as for a single neuron, at least the following neurophysiologic facts should be covered in an model of the discussed kind:

1. Each input signal causes some changes of the postsynaptic potential at the area of the input point; there are typical patterns of the increment and decay of the excitatory (EPSP) and inhibitory (IPSP) potential [1][2], 2. There is a set of commonly accepted neurophysiological data: time courses, amplitudes of various postsynaptic potentials, action-potential amplitude, and the thresholds for neural activation causing action potentials and removing the block of NMDA canals [1][3],

3. If (after an arrival of an action potential at a particular input) the accumulated local potential for postsynaptic region of each input synapse increase beyond the value of the local threshold (about -70mV), the NMDA receptor causes an opening of the previously blocked channel for an influx of calcium ions [4][5][6], and thus the phenomenon of Long–Term Potentiation (LTP), 4. On the most excitatory neural inputs the long-term memories can be activated, which results in an increase of LTP through an activation of NMDA receptors, 5. The amplitude of the single excitatory postsynaptic potential (EPSP) is not an absolute constant; indeed, several values has been proposed by neurophysiologists (see Andersen [3]), 6. Internal weights change due to propagation of postsynaptic potentials through the dendrite [7].

Some properties of the real neuron such as transiently after hyperpolarisation in refractory time and presynaptic regulatory mechanisms as reuptake of released neurotransmitters we did not take into consideration. As for currently available simulators of life-like neural circuits, they allow to simulate the circuits built of thousands of cells, but neglect some important properties of single cell (see Ambros & Ingerson [8]; Izhikevich [9] or tend to go deeply into detailed biophysics and biochemistry of a single neuron; so, in case of simulation of large neural networks, the complexity of the neuron's mathematical description requires hardly available computational power e.g. Hines [10]; Sikora [11]; Traub et al. [12]). It seems to be commonly believed in the research that a useful neural model must be of "channel-type" i.e. it must reflect opening of ionic channels at the cell membrane of dendrite as a reaction to action potential arrival at the synapse and opening ionic channels at the initial segment of axon for generation and propagation of action potentials (see Bower & Beman [13]). We propose that in neural simulations the computationally-expensive model of biophysical phenomena can be replaced with a model based on a set of three shift register and that no essential output properties are lost because of such replacement. The neuron properties we consider are frequency coding, a memory of a single input value, and a processing algorithm that facilitates a non-linear potential summation. Despite such a simplification, the proposed model demonstrates such phenomena as attenuation of receptors for external stimuli , delay and decay of postsynaptic potentials, modification of internal weights due to propagation of postsynaptic potentials through the dendrite, modification of properties of the analog memory for each input due to a pattern of LTP, outputspike generation when the sum of all inputs exceeds a threshold, and a refraction.

## 2. Firing Cell (FC)

The model of neuron we propose is called FC (firing cell) and occurs in three versions: excitatory, inhibitory and receptory (in this paper we discuss only the first one). FC consists of a dendrite, body and axon (Fig. 1). The dendrite is a string of compartments. Each compartment contains an input synapse. Each synapse is being checked and after a detection of an action potential the values of the table of a certain shift register are being changed according to the typical time course of EPSP or IPSP. Each of the registers is a string of data-boxes. The first data-box of one of the registers is reserved for the actual value of activation of the related compartment. Every 0,5 msec of the simulated time the actual values of all registers are checked and their weighted sums are compared with related thresholds. Generally, there is a positive correlation between the weight and a related synapse's proximity to the cell body. The output takes at a given clock the value equal to 1, if and only if the accumulated postsynaptic potential at the level of the neuron's body got at least as high as the threshold that equals -50mV. The axon provides the output signal to the compartments of the dendrites of the destination FCs. The switching signal is being back propagated to all dendrite compartments. The period between two state updates, i.e., between a given clock and the previous clock, is an equivalent of 0.5 msec of the simulated time. When an action-potential is generated, the registers are reset to the value of resting potential (using the switching signal) for and then the activity of all compartments is inhibited for 1.5 msec period of refraction. The values of the resting potential and threshold as well as other initial parameters can be input independently for each neuron before each simulation session. The minimal value of the single postsynaptic potential is calculated as dividing the difference between the threshold and the Kalium Equilibrium Potential -90mV by the number of inputs. For practical reasons, in first simulations we enhanced this value for EPSP by two. The object can then adjust all values proportionally to a given number of inputs.

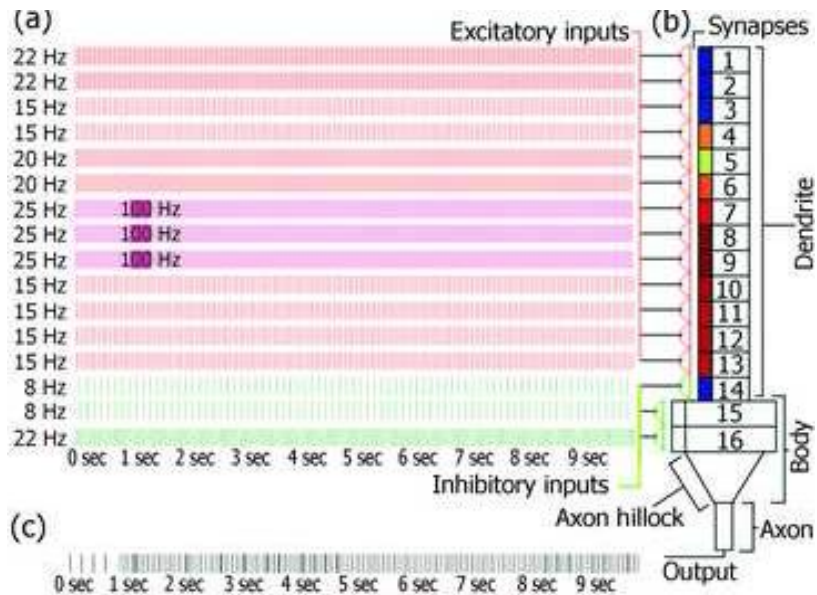

Fig. 1. Firing Cell (FC) in the configuration used in the experiments described in this paper.

We can consider the mentioned shift registers above as a very short-term local memory unit that remembers related events for up to 15 msec. The biological nervous system uses them to avoid the troublesome necessity of strict synchronization that takes pace in conventional computers. Yet we should also have a memory related to each input. The memory must work according to LTP rule. In each state-updating cycle the program calculates (based on actual values in the registers and appropriate weights) an accumulated local potential for postsynaptic region of each input synapse. If, after an action potential arrival at a particular input, the postsynaptic potential increase beyond the value of the local threshold (ca. -70mV), the second register (which simulates the rise of calcium ion content inside the cell) change and shift the contents of their data-boxes. Thus, the function operating on the registers substitute a time-consuming solving of differential equation related to the strength of synaptic potentiation versus calcium ion charge. The enhanced strength remains for a time calculated as power function of the charge. If after some time an additional charge appears, the period of the calculated synaptic potentiation substantially increases. In the same pattern we can easily include for each input the third register, which can simulate the slower influence of neuromodulatory substances on theirs receptors; however, in our preliminary investigations we did not use them.

## 3. Experiment

We used the simulated FC as a model of a pyramidal hippocampal cell under conditions such as in the classical LTP experiment by Bliss and Lomo [14]. We configured the FC in such a way that it had 13 excitatory inputs and 3 inhibitory inputs. Two cases were tested. In the first case the assumed EPSP amplitude was 5mV (Fig. 2), in the second it was 7 mV.

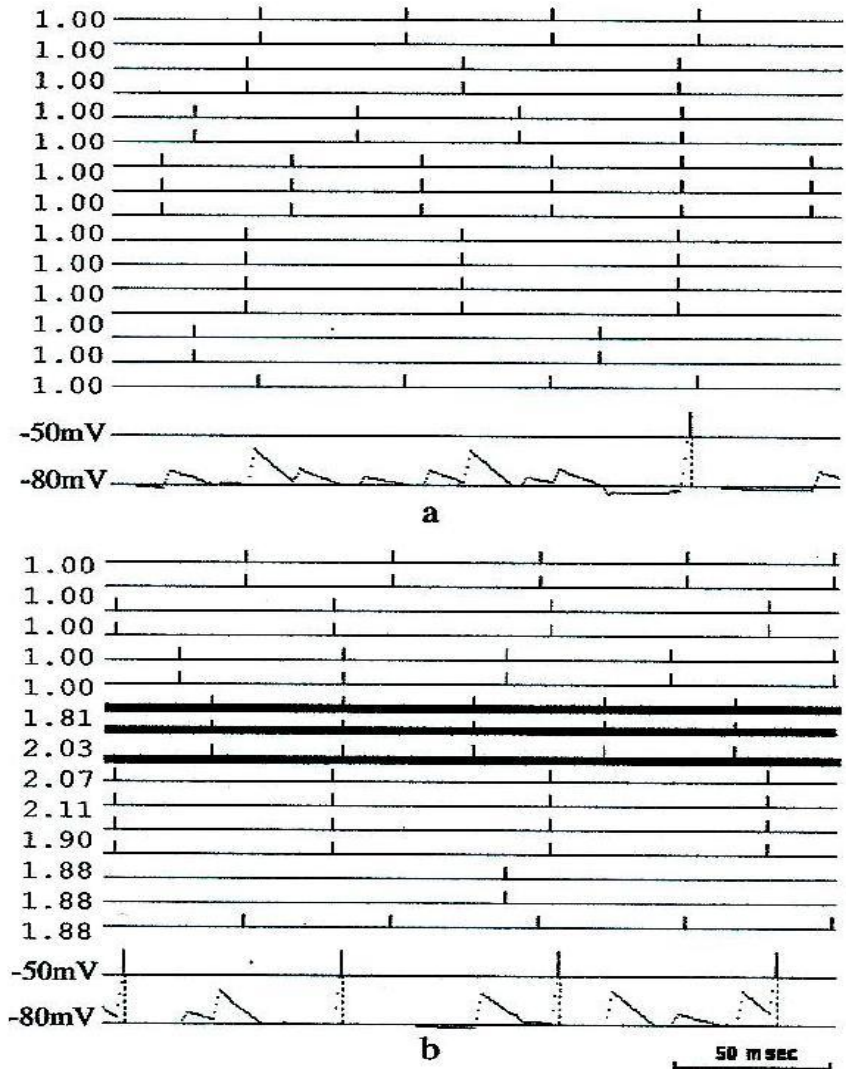

Fig 2. Behavior of an untrained FC (a) and trained FC (b) for EPSP 5mV. The peaks located on the input lines show history of action potential arrivals, while the vertical peaks above the threshold (-50mV) show action potential generation by FC. The zig-zag-shaped line between the threshold and resting potential (-80mV) is the plot of accumulated postsynaptic potential. The training of 400 msec at 100Hz was with action potential spike trains on the 7th, 8th and 9th input (b-thick lines). The frequency of action potentials and the state of the long term memory associated with the related input (real numbers on the left) increase after training (b). An animation of 10 sec of the firing cell's work is available at the web page: (http://medinf.gumed.edu.pl/383.html).

In the first case each FC's synapse substituted ca. 1000 biological synapses, whereas in the second case each FC's synapse substituted ca. 1400 biological synapses. In both cases three of the excitatory inputs were subject to a stimulation that lasted for 400msec with spike trains of the frequency 100Hz. The results of the experiment confirmed the biological plausibility of FC as for information processing- related mechanisms. In both tested cases the frequency of action-potential generation and the values of LTP increased after the training and in the case of EPSP amplitude 7mV the frequency were substantially higher both before and after the training. For demonstration of regular, periodic or chaotic behavior of a spiking neuron we used the Poincare return maps. Our experiments showed differences between the neuron's behavior before, during, and after the training. (Fig. 3) We also noticed that change of initial values implicate change of FC's behavior. For detailed description we need further statistical calculations with numerous simulations at various initial values.

a

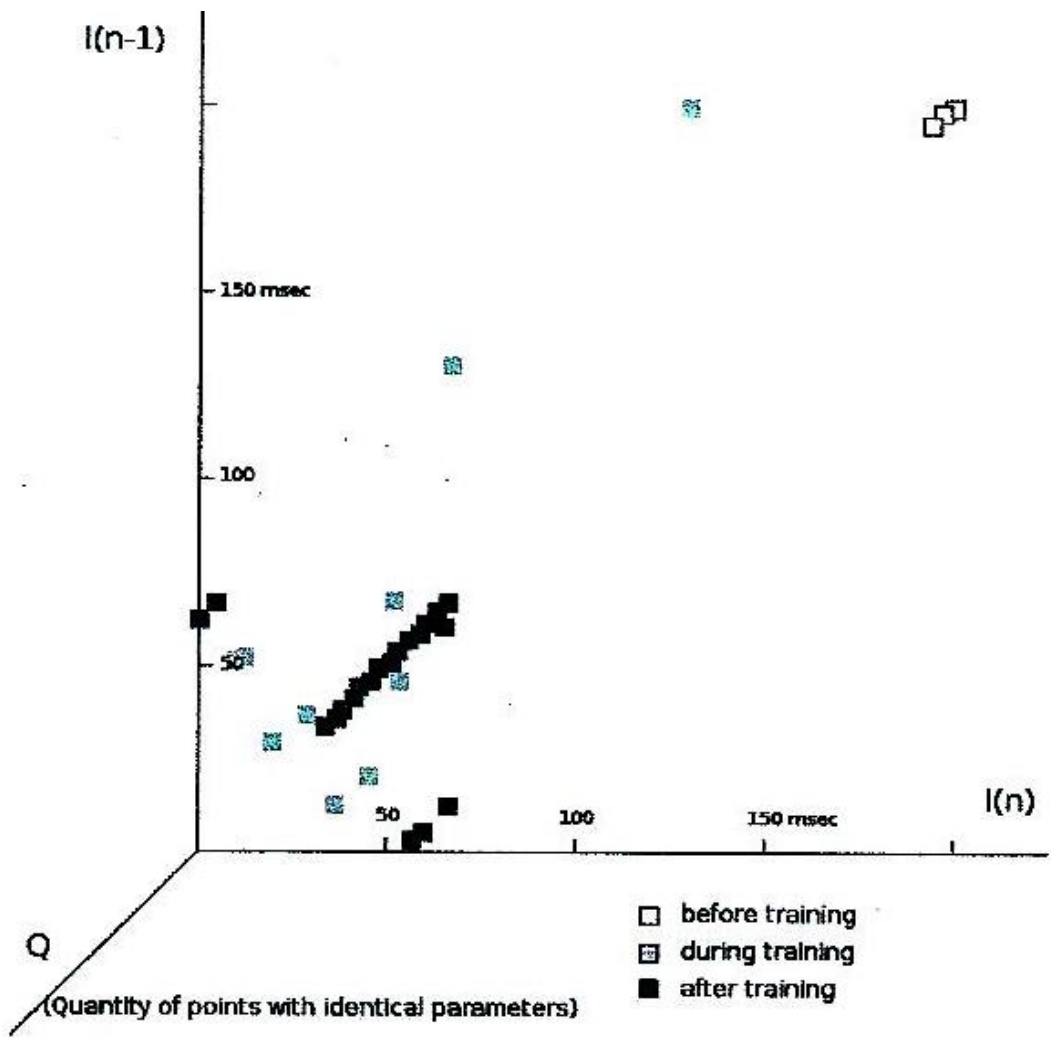

**b**

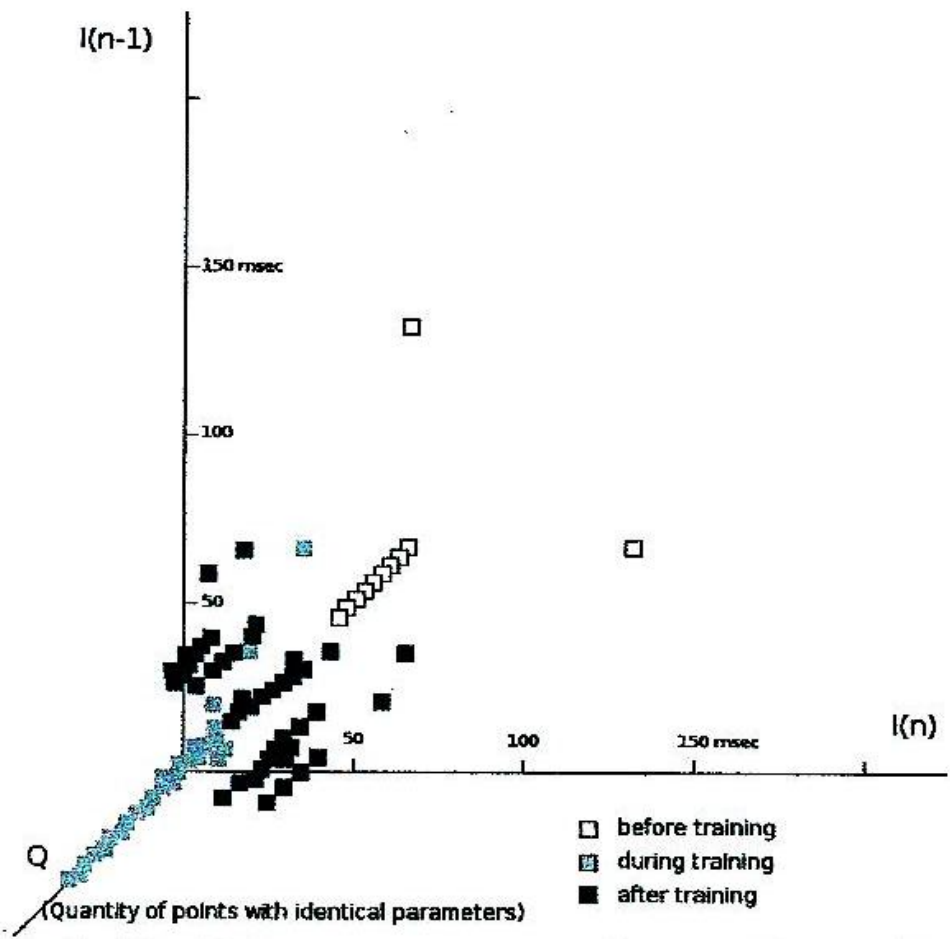

Fig 3. The Poincare return maps from total simulations time 2,5 sec. For EPSP 5mV (a) and 7mV (b). I(n) -interval between two subsequent spikes. Note the passages between various musters of spike trains.

### 4. Concluding Remarks

The way of information coding in the brain is still a mystery. Some neuroscientists suggest that encoding of information using firing rates is a very popular coding scheme used by cerebral cortex [15]. In the FC model this idea is followed thorough elaborating mainly the mechanisms that facilitate a frequency-based coding of information. To the best of our knowledge, no similar solutions have been published to date. The realistic modeling requires a lot of computational power. Note that Amaral & Ishizuka [16] calculated 12,000 synapses on a single rat's hippocampal neuron. The idea of representing near 1000 biological synapses by each of the FC's synapse is justified by the that during experiments in vivo a single electrode for technical reasons excites simultaneously multiple fibers. Hence, in information-processing oriented simulation there is no reason to calculate separately the states of thousands of synapses. The set of functions employed in the FC model should be a subject to further simplification toward a future implementation in hardware, for which analog memories and field transistors (as processing devices) are being considered. The silicon neuron by Mahovald and Douglas [17] seems to well solve the problem of the generation of action potentials. A satisfactory model must properly react frequency and phase of action-potential spike-trains. Some reported solutions e.g. Elias & Northmore [18] seem to be a good step in this direction.

**Acknowledgment:** This research was supported by Gdansk Artificial Brain Research Initiative (GABRI  http://www.gabri.pl ) and first published within The Proceedings of 11 International Symposium on Artificial Life and Robotics 2006 (AROB 11th'06, pp: 731-734) in Japan. In presented version only the fig. 1 is in enlarged version for technical reason, first author  actual contact address: jacekwb@gumed.edu.pl